\def\year{2022}\relax
\documentclass[letterpaper]{article} 
\usepackage{aaai22}  
\usepackage{times}  
\usepackage{helvet}  
\usepackage{courier}  
\usepackage[hyphens]{url}  
\usepackage{graphicx} 
\usepackage{natbib}  
\usepackage{caption} 
\DeclareCaptionStyle{ruled}{labelfont=normalfont,labelsep=colon,strut=off} 
\usepackage{algorithm}
\usepackage{xspace}
\usepackage{hyperref}
\usepackage{appendix}

\usepackage{newfloat}
\usepackage{listings}
\floatstyle{ruled}
\newfloat{listing}{tb}{lst}{}
\floatname{listing}{Listing}
\title{NICE: Robust Scheduling through Reinforcement Learning-Guided Integer Programming}
\author {
    Luke Kenworthy\textsuperscript{\rm 1},
    Siddharth Nayak\textsuperscript{\rm 2},
    Christopher Chin\textsuperscript{\rm 2},
    Hamsa Balakrishnan\textsuperscript{\rm 2}
}
\affiliations {
    \textsuperscript{\rm 1} US Air Force-MIT AI Accelerator\\
    \textsuperscript{\rm 2} MIT\\
    \{lukek, sidnayak, chychin, hamsa\}@mit.edu
}

\usepackage{amsmath}
\usepackage{multirow}
\usepackage{algpseudocode}

\newcommand{\name}{{NICE}\xspace}
\newcommand{\fullname}{{Neural network IP Coefficient Extraction}\xspace}

\begin{document}

\maketitle

\begin{abstract}
    Integer programs provide a powerful abstraction for representing a wide range of real-world scheduling problems. Despite their ability to model general scheduling problems, solving large-scale integer programs (IP) remains a computational challenge in practice. The incorporation of more complex objectives such as robustness to disruptions further exacerbates the computational challenge. We present \name (\fullname), a novel technique that combines reinforcement learning and integer programming to tackle the problem of robust scheduling. More specifically, \name uses reinforcement learning to approximately represent complex objectives in an integer programming formulation.  We use \name to determine assignments of pilots to a flight crew schedule so as to reduce the impact of disruptions. We compare \name with (1) a baseline integer programming formulation that produces a feasible crew schedule, and (2) a robust integer programming formulation that explicitly tries to minimize the impact of disruptions. Our experiments show that, across a variety of scenarios, \name produces schedules resulting in 33\% to 48\% fewer disruptions than the baseline formulation. Moreover, in more severely constrained scheduling scenarios in which the robust integer program fails to produce a schedule within 90 minutes, \name is able to build robust schedules in less than 2 seconds on average.
    
\end{abstract}

\section{Introduction}
    Scheduling is a ubiquitous type of optimization problem that is often solved using integer programs (IPs) or mixed-integer programs (MIPs). While IPs and MIPs can model a wide range of scheduling problems, solving large-scale instances in practice can be computational challenging. 
    
    In most practical applications, it is also important that the schedules are robust to uncertainties. Robust scheduling involves the building of schedules that will undergo minimal change when faced with unknown future disruptions. While MIP formulations of scheduling problems can be extended to account for robustness, the resulting problems are often much more computationally challenging than their baseline, non-robust counterparts. Even with state-of-the-art solvers, such extensions to accommodate robustness can add hours to the time needed to compute an optimal schedule, sometimes making them impractical for real-world use.
    
    We propose a technique, \fullname (\name), that seeks to find a quick-but-approximate solution to a scheduling problem with an additional robustness objective, by using reinforcement learning (RL) to guide the IP formulation. We first determine a \emph{feasible} schedule using a baseline IP. We simultaneously train an RL model to build a schedule for the same problem, using a reward function that leads to more robust schedules (but that would have added considerable computational burden if encoded directly in the IP formulation). Then, rather than use the RL model to create a schedule directly, we use the probabilities in its output layer to assign coefficient weights to the decision variables in our simpler IP to create a feasible schedule. By doing so, we leverage the intuition behind knowledge distillation \cite{hinton2015distilling} that the distribution of values in the output layer of a neural network contains valuable information about the problem. 
    
    \name allows us to approximate the robust scheduling
    formulation with significantly fewer variables and constraints. Across a variety of disruption scenarios, we find that \name creates schedules with 33--48\% fewer changes than the baseline. Moreover, in certain practical problem instances, \name finds a solution in a matter of seconds; the corresponding IP that explicitly optimizes for robustness fails to produce a solution within 90 minutes for the same scenarios.
    
    The main contribution of this paper is the introduction of a new technique to approximate complicated IP formulations using RL. To the best of our knowledge, \name\footnote{Our code is available at \href{https://github.com/nsidn98/NICE}{https://github.com/nsidn98/NICE}} is the first method to use information extracted from neural networks in IP construction. We illustrate the performance of \name in creating robust (disruption-resistant) crew schedules (i.e., assignments of pilots to flights). However, robust crew scheduling is only one application of \name; we believe that the method is potentially applicable to a wider range of discrete optimization problems.

\section{Background and Related Work} \label{background}

    Personnel scheduling has been a long-standing challenge in Operations Research, and has been the focus of much research over the past several decades \cite{Dantzig1954, VanDenBergh2013}. Integer programs (IPs) and mixed integer programs (MIPs)\footnote{We use the terms integer program (IP) and integer linear program (ILP) interchangeably unless noted otherwise; similarly for mixed-integer program (MIP) and mixed-integer linear program (MILP).} have been widely-used for personnel scheduling, in large part due to their ability to represent general scheduling problems. However, despite the power of IPs to model scheduling problems, solving large-scale IPs in practice is often  computationally challenging \cite{Papadimitriou1998}. Most real-world applications also need robust schedules, namely, schedules that do not require considerable adjustments to personnel assignments in the event of an unforeseen disruption. While robust MIP-based formulations of scheduling problems can be developed, they are usually at least as computationally challenging as their non-robust counterparts \cite{van2017robust, vujanic2016robust,Craparo2017, BertsimasSim2003}.   
    
    The scheduling of flight crews (e.g., pilots) is a personnel scheduling problem that arises in the context of aviation \cite{Honour1975,Caprara1998, Navy2, Zhang2020-ACRP}. Similar to other scheduling problems, crew scheduling has traditionally been tackled using large-scale IPs, both for airline and military flight crews \cite{Kohl2004,TOPGUN}
    Flight delays are the main cause of disruption to crew schedules in commercial aviation;  buffers (or slack in the schedule) have therefore been considered as a mechanism to achieve schedule stability amidst flight delays \cite{buffer1, buffer2}. A \emph{buffer} refers to the amount of time between the successive flights flown by a particular pilot. By increasing these buffers, a new pilot assignment is less likely to be needed due the initially-planned pilot being delayed, and therefore being unable to make the flight. 

    \subsubsection{Learning-Based Approaches to Scheduling} Successes in deep (reinforcement) learning have motivated research that focuses on obtaining end-to-end solutions to combinatorial optimization problems; e.g., the travelling salesman problem \cite{vrpRL1, vrpRL2, vrpRL3, vrpRL4, vrpRL5} or the satisfiability problem (SAT) \cite{SAT1, SAT2}. RL has also been used for resource management and scheduling in a diverse set of real-world applications. For example, Gomes \shortcite{RLHealthcareScheduling2017} uses an asynchronous variation of the actor-critic method (A3C) \cite{A3C} to minimize the waiting times of patients at healthcare clinics. The lack of readily available optimization methods for this problem due to the ad-hoc nature of patient appointment scheduling motivates the usage of RL for this application to model the uncertainty. Mao et al. \shortcite{DeepRM} introduce DeepRM which uses resource occupancy status in the form of images as the states and uses neural networks for training the RL agent. Chen et al. \shortcite{multiJobSchedulingRL2018} improve upon DeepRM by modifying the state-space, reward structure and the network used in the DeepRM paper. Chinchali et al. \shortcite{ChinchaliNetworkTraffic_2018} use RL for cellular network traffic scheduling. They incorporate a history of the states observed to re-cast the problem as a Markov Decision Process (MDP) \cite{MDP} from a non-Markovian setting. Along with this modification, they construct a reward function that can be modified according to user preferences. In all of the works above, the models used application-specific state space, action space, and reward structures to optimize special-purpose objective functions to create schedules.

    In recent work, Nair et al. \shortcite{MIPwRL} use a bipartite graph representation of a MIP and leverage graph neural networks \cite{gnn, gcn} to train a generative model over assignments of the MIP's integer variables. We do not use generative models to solve an IP formulation, but instead use RL to formulate the IP itself. Very recent work by Ichnowski et al. \shortcite{RLQP} uses RL to speed up the convergence rates in quadratic optimization problems by tuning the inner parameters of the solver.

    \subsubsection{Knowledge Distillation} Hinton et al. \shortcite{hinton2015distilling} explored the use of internal neural network values to distill the knowledge learned by a model. They reasoned that the probabilities in a neural network's output layer carry useful information, even if only the maximum probability value is used for ultimate classification: ``An image of a BMW, for example, may only have a very small chance of being mistaken for a garbage truck, but that mistake is still many times more probable than mistaking it for a carrot.'' In this work, they used the values from the input to the output layer of a larger neural network, as well as the training data itself, to train a smaller neural network. This smaller neural network achieved fewer classification errors than a network of the same size trained only on the training data. Using a similar approach, they trained a neural network on speech recognition data with the same architecture as a neural network trained on the data directly. They found that the new, distilled model performed better than the original one; it also matched 80\% of the accuracy gains attained by averaging an ensemble of 10 neural networks with the same architecture, each initialized with different random weights at the beginning of training. Similarly, our approach uses the probabilities output by a neural network to extract objective function coefficient weights. 
    
\section{Approach} \label{sec:approach}

    \subsection{Crew Scheduling Problem}
    
    In this paper, we seek to build robust schedules for a version of the flight crew scheduling problem, hereafter referred to as the ``crew scheduling problem,'' which has a long history in both commercial and military aviation \cite{arabeyre1969airline, gopalakrishnan2005airline, combs2004hybrid}. We consider the scenario in which we are given a collection of flights that must be flown by a given squadron (i.e., a group) of pilots. Every flight has multiple slots, each of which must be filled by a different pilot. Each slot has qualification requirements that must be satisfied by any pilot who is assigned to that slot. Depending on their qualification, a pilot would only be eligible to fill a subset of slots. Finally, every pilot has some specified availability.
    We discretize our schedule into days, although other time discretizations could be used. In other words, a feasible schedule assigns pilots to slots such that every flight in the schedule horizon is fully covered, and the qualification requirements for the slots and availability restrictions of the pilots are satisfied. 
    
    We worked with a flying squadron to develop our problem formulation, focusing on their constraints and preferences.  Consequently, our formulation differs from some of the crew scheduling formulations in prior literature, which have been largely in the context of airline flight crews. For example, all of the squadron's flights start and end at the same place, so we do not factor in crew relocation. Also, in the data we received, the start and end dates of the flights included the required crew rest time, so we did not need to explicitly model this. However, flying squadrons have more granular pilot qualification levels than have been considered in airline crew scheduling. 
    
    \subsection{Baseline Integer Program Formulation}
    Chin \shortcite{Chin2021} and Koch \shortcite{Koch2021} created a baseline IP for the crew scheduling problem, producing a satisfactory assignment with respect to all relevant constraints. We use a similar construction for our baseline IP, with the primary difference of 
    using a decision variable for the assignment of each pilot to each slot, rather than one for each pilot to each flight. 
    We define the following sets and subsets:
    {\small
    \begin{equation*}
    \begin{split}
        i \in I & \text{\quad The set of pilots} \\
        f \in F & \text{\quad The set of flights} \\
        s \in S & \text{\quad The set of all slots} \\
        U_f \subset F & \text{\quad Flights that conflict with flight $f$} \\
        L_i \subset F & \text{\quad Flights that conflict with pilot $i$'s leave} \\
        S_f \subset S & \text{\quad The set of slots belonging to flight $f$} \\
        Q_i \subset S & \text{\quad Slots that pilot $i$ qualifies for} \\
    \end{split}
    \end{equation*}}
    
    We use the binary decision variable $X_{is}$, which is 1 if pilot $i \in I$ is assigned to slot $s \in S$, and 0 otherwise. We now have the following equation:

    \begin{equation}
        \max \sum_{i \in I} \sum_{s \in S} X_{is} \tag{1.1}\label{eq:1.1}
    \end{equation}
    
    such that:
    {\small
    \begin{subequations}
    \begin{align}
        X_{is} = 0 && \forall i \in I, f \in L_i, s \in S_f \tag{1.2}\label{eq:1.2} \\
        X_{is} = 0 && \forall i \in I, s \in S \setminus Q_i \tag{1.3}\label{eq:1.3} \\
        \sum_{s \in S_f} X_{is} \leq 1 && \forall i \in I, f \in F \tag{1.4}\label{eq:1.4} \\
        \sum_{i \in I} X_{is} = 1 && \forall s \in S \tag{1.5}\label{eq:1.5} \\
        \sum_{s \in S_f} X_{is} + \sum_{s' \in S_{f'}} X_{is'} \leq 1 && \forall i \in I, f \in F, f' \in U_f \tag{1.6}\label{eq:1.6} \\
        X_{is} \in \{0, 1\} && \forall i \in I, s \in S  \tag{1.7}\label{eq:1.7}
    \end{align}
    \end{subequations}}
    
    Constraints \eqref{eq:1.2} and \eqref{eq:1.3} ensure that pilots are only assigned to slots that they are qualified for, and that pilots will never be assigned to flights that conflict with their leave. Constraint \eqref{eq:1.4} prevents the scheduling of pilots to multiple slots on the same flight. Constraint \eqref{eq:1.5} ensures that every slot gets filled by exactly one pilot. Constraint \eqref{eq:1.6} prevents assignment of pilots to conflicting flights. We make our pilot-slot decision variable binary with Constraint \eqref{eq:1.7}. Finally, Equation \eqref{eq:1.1} ensures that we fill as many slots as possible. Due to our requirement that each slot have exactly one pilot associated with it, this equation always produces the same objective value.

    \subsection{Buffer Formulation} 
    
    Chin \shortcite{Chin2021} utilizes additional constraints and decision variables to optimize schedules for robustness by increasing the amount of buffer time between flights. For our purposes, we consider the buffer time between two flights to be the number of full days between their start and end time. For example, suppose pilot \textit{X} is assigned to flights \textit{A} and \textit{B}; flight \textit{A} ends on day 1, and flight \textit{B} starts on day 5. We say that there is a buffer time of 3 days for pilot \textit{X}.
    
    To incorporate buffers into an IP, Chin \shortcite{Chin2021} identifies 
    all flight pairings that would create a buffer less than or equal to some maximum threshold, $T_{\text{buffer}}$, used to dampen the complexity of realistic problem instances. Then, 
    \{0, 1\} decision variables $B_{iff'}$ for all pilots $i \in I$ and flights $f, f' \in F \times F$ are created. 
    Constraints are used to ensure that $B_{iff'}$ is $1$ if and only if the following conditions are met:
    
    \begin{enumerate}
        \item $f \neq f'$
        \item Pilot $i$ qualifies for at least one slot in both $f$ and $f'$.
        \item $f$ and $f'$ have a buffer between $0$ and $T_{\text{buffer}}$, inclusive.
         \item Pilot $i$ is assigned to consecutive flights $f$ and $f'$.
    \end{enumerate} 
    
    Chin \shortcite{Chin2021}
    then defines a buffer penalty $b_{iff'} \in [-1, 0)$ that gets more negative for lower buffers: down to -1 when the buffer between $f$ and $f'$ is 0, and closest to 0 when the buffer is $T_{\text{buffer}}$. Note that buffers longer than $T_{\text{buffer}}$ effectively have a penalty of 0. Finally, 
    he 
    incorporates all of this into the following objective function to optimize buffer time:
    
    \begin{equation}
        \max{} \sum_{i \in I} \sum_{f, f' \in F \times F} b_{iff'} B_{iff'}
    \end{equation}
    
    In our buffer IP, we incorporate this formulation with the assistance of additional auxiliary \{0,1\} decision variables for each pilot-flight combination, setting it to 1 if a pilot is assigned to any slot on a given flight, and 0 otherwise. In our experiments, we found that, in certain problem instances, the buffer IP was highly effective in producing robust schedules. 
    
    \subsection{Reinforcement Learning Formulation}
        We model the building of a valid schedule with a discrete event simulation (DES): at each time step, we take an action, which affects the state of our system. The main idea of our reinforcement learning scheduling approach is to order the slots that need to be scheduled and, at each slot, pick a pilot to assign to that slot. This technique, previously used by Washneck et al. \shortcite{Waschneck} for production scheduling, gives us an action space size equal to the number of pilots. If we get through all of the slots, we end up with a complete schedule, though filling all of the slots is not a guarantee; the RL scheduling agent could back itself into a corner, leaving no pilots to assign to a given slot based on its previous decisions. We use Proximal Policy Optimization (PPO) \cite{PPO}, which is an actor-critic method where the actor chooses the action for the agent and the critic estimates the value function. The actor network gives a probability distribution over the pilots to choose given the state input and the action is chosen by sampling from this distribution.

    \subsection{\name}

        \subsubsection{Motivation}
            In our early exploration of building schedules with non-robustness objectives, the RL-produced schedules would often approach the effectiveness of the IP schedules with respect to the optimized metrics, but they would never do better. This observation gave us two key insights. 

            First, the RL model assigns pilots to slots sequentially, operating in a ``greedy'' fashion. Once an assignment is made, it cannot be changed. Thus, the RL agent lacks a global view of the full schedule in its state space, meaning it has imperfect information at the time of each pilot assignment. 
            In contrast, the IP scheduling approach, which optimizes an objective function across all pilot assignments, can factor in tradeoffs created by the complex interplay of related constraints. 

            Second, it was clear that our RL scheduling agent was capable of learning. While it could not match the performance of the IP schedules in our preliminary exploration, it still produced schedules with considerably better objective performance than the baseline. When scheduling objectives can easily be captured in an IP, this observation is not particularly helpful. However, for more complicated optimizations that integer programming struggles with, this insight proves useful: to avoid the greedy pitfalls of RL for scheduling while still leveraging the knowledge learned by our neural network, we can use the probabilities produced by the output layer in our IP scheduling formulation.

            These two observations motivated the creation of \name. \name uses RL to approximate sophisticated integer programs with a simpler formulation. We apply this technique 
            to the crew scheduling problem.

        \subsubsection{\name IP Formulation} 
            As an IP, \name closely resembles the baseline formulation. The only difference is in the coefficients for the objective function.
            Recall that our decision variable, $X_{is}$, is 1 if pilot $i$ is assigned to slot $s$ and 0 otherwise. This variable aligns neatly with our RL scheduling formulation, which considers slots in a fixed order. At each slot, it produces a probability for each pilot that captures how likely assigning that pilot to that slot is to maximize reward in a given scheduling episode. It then assigns the pilot with the maximum probability to that slot. We can use $a_{is}$ to refer to the probabilities output by the network for the assignment of pilot $i$ to slot $s$. Now, to leverage the knowledge learned by the RL scheduling approach in our IP formulation, we can incorporate $a_{is}$ into our objective function:
            \begin{equation}
                \label{eq:3}
                \max \sum_{i \in I} \sum_{s \in S} a_{is} X_{is}
            \end{equation}
            With this new equation, our IP scheduling approach is incentivized to pick the pilot with the highest probability  possible at each slot, subject to constraints and possible rewards for other slots. 
            This new formulation approximately captures the reward function used by the RL agent while giving it a global view of pilot assignment.

        \subsubsection{Extracting Probability Weights}

            An important issue we had to address was the extraction of $a_{is}$ from our RL neural network. The actions taken at each state of the RL scheduling process can impact the pilot probability vector at later states. Thus, the order that the RL scheduler fills the slots has a potentially confounding impact on the $a_{is}$ weights. Extracting the probability vector at each slot while running the RL scheduling process as normal could cause the specific actions taken to bias our $a_{is}$ values, diminishing the advantage of the IP scheduler's global outlook. 

            In our experiments, we used two different approaches. In the first one, we took a Monte Carlo approach, trying to approximate the average weight across all possible orders of scheduling the slots. In this approach, we randomly shuffled the order of slots that the RL scheduler had to assign and then recorded the $a_{is}$ weights at each step of the scheduling process. We ran this process $n$ times to get $n$ total $a_{is}$ values for each pilot-slot pair. If the RL agent could not fill a slot in that round of scheduling, we set the $a_{is}$ value to 0. We then averaged the $n$ values for each pilot-slot pair to get our final $a_{is}$ values. Note that this method causes the scheduling process to take longer for higher values of $n$ because it has to run more RL scheduling rounds. 

            The next approach, which we call the ``blank slate'' approach, exploits the fact that the probability weights for the pilots produced by the first slot do not depend on any previous actions taken. Thus, we can make each slot our first scheduled slot to get weights that do not depend on previous decisions. To do so, for each slot $s$ in our fixed order, we initialized a new RL agent with the same underlying neural network, cutting out all of the states that occur before $s$ then extracting the $a_{is}$ values for the pilots on that slot. 

\begin{figure}[t]
            \centering
            \includegraphics[width=0.8\linewidth]{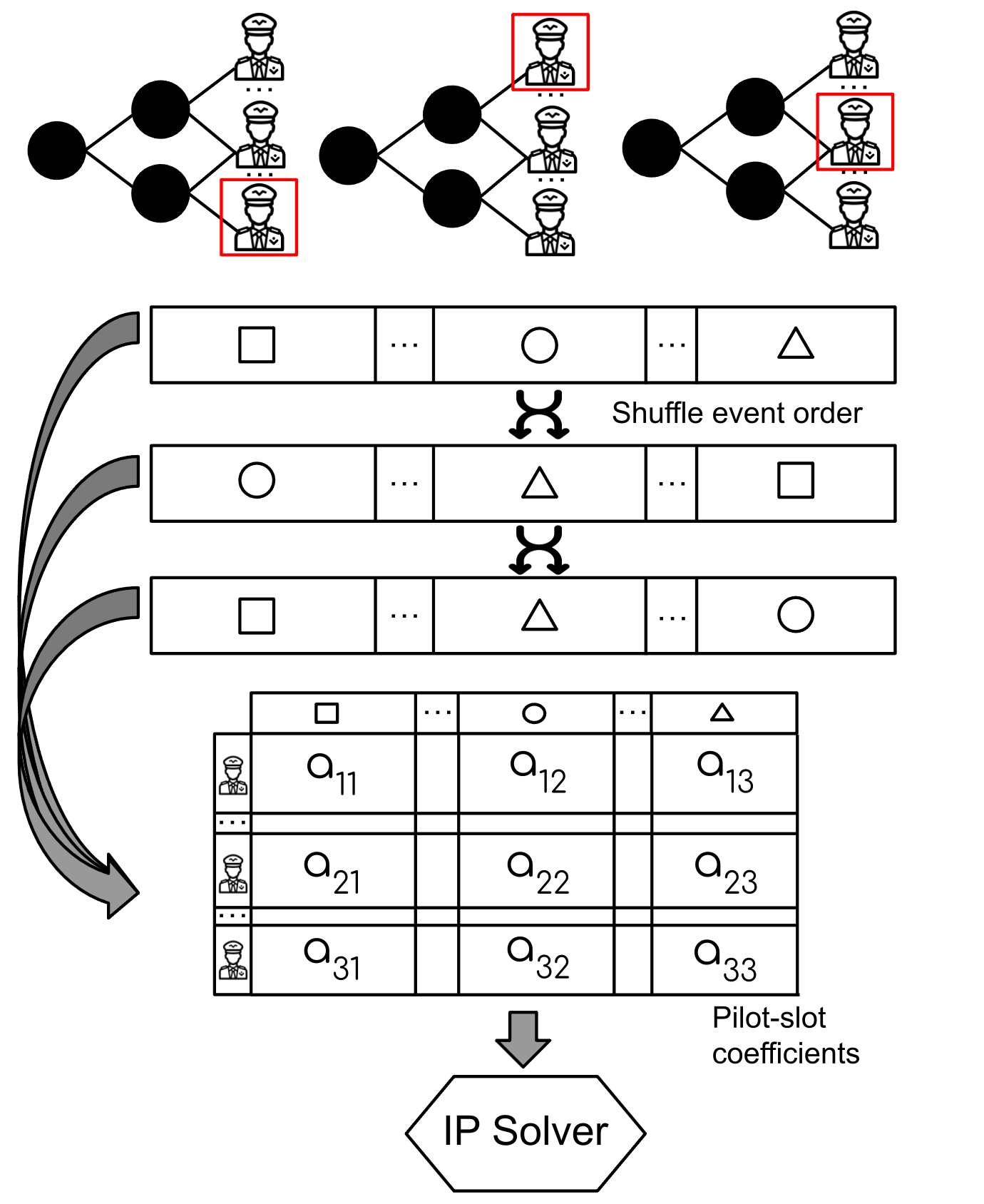}
            \caption{Once we train our network, in the Monte Carlo approach, we run the scheduler $n$ times, shuffling the order of slots each time. We then average the probability values in the output layer across the $n$ runs over pilots and slots to obtain our \name coefficients. Finally, we pass these coefficients to the IP solver to obtain our \name-generated schedule.}
            \label{fig:extract}
        \end{figure}

        Figure \ref{fig:extract} demonstrates using the Monte Carlo approach to extract weights from the neural network, then feeding those weights into the IP scheduler. To summarize, to build our NICE schedule:

        \begin{enumerate}
            \item We train a neural network on the DES version of the crew scheduling problem.
            \item We use either the Monte Carlo or ``blank slate'' approach to extract probabilities from the output layer of the neural network for the assignment of each pilot to each slot.
            \item For each pilot-slot combination, we use the extracted probability as the coefficient ($a_{is}$) for its respective pilot-slot decision variable ($X_{is}$) 
            \item Using this objective function and its constraints, we solve the IP to obtain our scheduling solution.
        \end{enumerate}

        We note that while we apply NICE specifically to the crew scheduling problem in this paper, \name is highly generalizable as it can be used to obtain a solution to any problem where there is an isomorphism between the IP and DES formulation.

    \subsubsection{Incorporating Robustness}
    
        To use \name to build robust schedules, we train an RL scheduler to optimize buffer time in its assignments. To do so, we include a reward of $b + 1$ whenever the RL agent places a pilot on an event that forms a buffer of length $b$ with the pilot's most recent event. We add the $+ 1$ term to reward the agent for making a placement, regardless of buffer. We give the agent a reward of $T-1$ when it places a pilot with no previous events scheduled. We chose $T-1$ because this is the maximum reward for the $T$-day schedules that the RL agent trains on. For example, imagine pilot \textit{X} is assigned to a 0-day flight starting and ending on day 1. If the pilot were assigned to a flight starting on day 7, that assignment would earn a reward of $6$, because there are 5 days between day 1 and day 7, exclusive. We do not use a maximum buffer value like $T_{\text{buffer}}$ in the IP formulation because larger $T_{\text{buffer}}$ values do not noticeably affect the run time of our program like it does with the IP. We included two exceptions to this reward policy. First, to incentivize building full schedules, the agent earns a reward of $25$ when all slots in an episode are scheduled. Second, to deter the creation of incomplete schedules, we give the agent a reward of $-10$ when it is unable to schedule all events in an episode. 
    
        In short, we give local rewards to our RL agent to help it build complete, robust schedules using buffers as a heuristic; our hope is that this training method will ultimately produce probability weights that, when extracted, lead our IP solver to build robust schedules.

\section{Experiments}
    \label{sec:random_event_gen}
    \subsection{RL Training}
        To train our RL agent, we created an OpenAI Gym \cite{gym} type environment for the crew scheduling problem. We utilized an anonymized dataset from a 
        flying squadron to construct a random event generator. The dataset contained 87 pilots with 32 different qualifications and 801 flights across over six months, each containing between 2--3 slots. There were 16 different types of flights, where the type determined the qualification requirements for the slots on the flight. These flights were subdivided into two categories: missions and simulators, which we treat equivalently except for the purposes of our random event generation. There were 7 mission types and 9 simulator types. We trained our RL agent on randomly generated flights based on this dataset. 

    \subsubsection{Random Event Generation}
        To generate the random flights, we first divided our dataset into simulators and missions that started in 26 different full-week intervals. For each week, we created $\alpha$ random mission-based flights, where $\alpha$ is drawn from a normal distribution with a mean and standard deviation equal to the mean and standard deviation of missions across all 26 weeks. Similarly, we also create $\beta$ random simulator-based flights, where $\beta$ is drawn from a similar distribution that uses simulators instead of missions. The dataset also contained a variety of training requirements that each pilot, ideally, would fulfill; the number of times the pilot should fulfill each requirement; and information about which flight satisfied which training requirements. Along with the training requirement information, each flight contained two binary training requirement qualifiers (TRQs) to help determine which training requirements the flight fulfilled. Note that we did not use the training requirement information outside of the state space formulation for our RL scheduler. In our experiments, we found that including the training requirements from our dataset in the state space helped our NICE scheduler perform better. We suspect that they helped our neural network better reason about trade-offs when selecting a particular pilot for a slot.

        For each flight generated, we picked a random day in the scheduling week for it to start. We then randomly pick a type for the flight, where the probability of picking that type of mission was proportional to the number of times it showed up in the dataset. To determine the length of duration for the flight, we randomly sample a flight length from a flight of the selected type from the dataset. 
        We followed an identical process to generate each simulator, except each simulator started and ended on the same day, so we did not randomly pick a length. The dataset provides the dates each pilot is on leave, which we used directly. Then, we created a fixed ordering of slots that need to be scheduled. We order the slots first by the corresponding flight's start date and use the flight's arbitrary unique ID as a tie-breaker. We order slots of the same flight by ascending qualification. The assignment of a pilot to a slot serves as a time-step in our DES. 

    \subsubsection{Configuration}
        The state space for the RL agent includes:
        \begin{enumerate}
            \item A binary vector for the pilots available for the current slot
            \item A flattened vector encoding the current event to be scheduled, consisting of:
            \begin{enumerate}
                \item A one-hot encoding of the event type
                \item A binary vector indicating whether each TRQ was true or false
                \item A binary vector representing the pilots assigned to the current event. (Each event may have 2-3 pilot slots)
                \item The event duration (in days)
                \item The number of days between the start of the scheduling episode and the start of the event
                \item The number of days between the start of the scheduling episode and the end of the event
            \end{enumerate}
            \item A vector containing the total number of training requirement fulfillments each pilot could receive for flying that event, if it were flown a sufficient number of times
        \end{enumerate}

    To build the neural network for our \name scheduler, we trained a variety of models with different hyperparameter combinations. One of the hyperparameters of particular interest was the training schedule density, $d$. Recall that we scheduled $\alpha$ flights and $\beta$ simulators in a round of scheduling. During training, we multiplied $\alpha$ and $\beta$ by $d$ so the agent would schedule more flights in the same time period. We trained RL models with $d \in \{1, 2, 3\}$. We trained a model with 5 different seeds for each value of $d$, ultimately creating $3 \times 5 = 15$ different neural networks for testing. The hyperparameters for all our experiments are listed in the Appendix \cite{NICEarxiv}. We note that, because the output layer is equal to the size of the number of pilots, adding a pilot would require us to re-train the network with the new shape.

    \begin{table*}[t] \centering
        \begin{tabular}{|c|c|c|c|c|r|c|c|}
        \hline
            {\% Flights} & \multicolumn{4}{c|}{Number of Disruptions} & \multicolumn{3}{c|}{Significance of Difference ($p$-value)} \\ \cline{2-8} 
            Delayed   & \name           & Baseline IP     & RL              & Buffer IP       & \name-Baseline& \name-RL  & \name-Buffer        \\ \hline
            25                                  & $\textbf{0.34}\pm \textbf{0.71}$  & $0.61 \pm 1.07$ & $32.6 \pm 7.33$ & $0 \pm 0$ & 0.03~~~~~       & $<0.01$   & $<0.01$             \\ \hline
            50                                  & $\textbf{0.67} \pm \textbf{0.92}$ & $1.16 \pm 1.55$ & $27.1 \pm 7.13$ & $0 \pm 0$ & $<0.01$~~~~~   & $<0.01$  & $<0.01$        \\ \hline
            75                                  & $\textbf{0.66} \pm \textbf{0.99}$ & $1.13 \pm 1.73$ & $23.0 \pm 6.34$ & $0 \pm 0$ &0.01~~~~~        & $<0.01$  &  $<0.01$        \\ \hline
            100                                 & $\textbf{0.63} \pm \textbf{0.82}$ & $1.06 \pm 1.50$ & $17.9 \pm 6.29$& $0 \pm 0$ &0.01~~~~~        & $<0.01$  & $<0.01$            \\ \hline
        \end{tabular}
        \caption{Average and standard deviation of disruptions across scheduling methods when flights are delayed (lower values are better). Scheduling density of 1. }
        \label{disrupted_results_pm}
    \end{table*}

    \subsection{Scheduling Parameter Selection}

        For further experimentation, we had to pick the best combination of RL model and weight extraction method to use. We parameterize our weight extraction methods with the variable $n$, where $n = 0$ represents the ``blank slate'' extraction method mentioned previously, and $n > 0$ represents the $n$ value used in the Monte Carlo approach. 

        To select the best combination, we built an environment where, using the same flight generation process to train the RL scheduler, we generate 1 week's worth of flights. From these flights and associated slots, we generate pilot-slot pairings using the baseline integer program and the \name schedule with weights extracted from the selected RL network and the given $n$ value. Next, one day into the schedule, we delayed 50\% of the flights that had not already left. We pushed back each flight by a number of days randomly chosen uniformly between 1 and 3, figuring that delays longer than 3 days were relatively rare. To fix this disruption in both of these schedules, we used an integer program that minimized the number of changes to the pilot-slot pairings.
        From this disruption resolver, we end up with the number of disruptions that occurred under each schedule. Sometimes, we would randomly generate a series of flights that made it impossible for the IP or \name approach to schedule because a pilot-slot pairing did not exist that met all of the constraints. In these cases, we skipped to the next series of randomly-generated flights, not recording any disruption data because there was no schedule to disrupt. Then, for each neural network and for each $n$ value $n \in \{0, 2, 4, 8\}$ (60 experiments total), we ran this scenario 20 times, comparing the average number of schedule disruptions between the baseline IP and the \name scheduler across those 20 runs.

        We calculated the ratio of disruptions, $r$, between the two methods, where $r < 1$ indicates that the \name method provided a schedule with fewer disruptions  on average than the IP baseline. We then determined the median $r$ value over seed values for each training density and weight extraction method combination. We used the lowest median $r$ value to determine our final scheduling method and underlying neural network for \name. As a result of this process, we obtained a network trained with a density of 1 and an $n$ value of 2. In this case, across seed values, three networks produced the same median value for this combination, so we arbitrarily chose a network from those three models. The median $r$ value was $0.25$, with a range of 0.08 (0.25 to 0.33). We used this network and $n$ value in our further experiments. 

        Overall, the combinations had fairly stable performance over seed values, with the highest range in $r$ across seeds being $0.81$. Notably, the $n$ value of 0 produced the 3 highest ranges (0.81, 0.69, and 0.56), indicating that the ``blank slate'' weight extraction method led to a wider range of disruptions in produced schedules across random seeds. 

        \begin{table*}[t]
            \centering
            \begin{tabular}{|c|c|c|c|c|c|}
                \hline
                {\% Flights} & \multicolumn{3}{c|}{Number of Disruptions} &  \multicolumn{2}{c|}{Significance of Difference ($p$-value)} \\ \cline{2-6} 
                 Delayed   & \name           & Baseline IP     & RL               & \name-Baseline & \name-RL \\ \hline
                25  & $\textbf{1.99} \pm \textbf{1.99}$ & $2.99 \pm 2.68$ & $59.2 \pm 13.1$  & $<0.01$ & $<0.01$ \\ \hline
                50  & $\textbf{3.17} \pm \textbf{2.27}$ & $5.01 \pm 4.33$ & $51.2 \pm 12.2$  & $<0.01$ & $<0.01$ \\ \hline
                75  & $\textbf{3.32} \pm \textbf{2.33}$ & $6.32 \pm 5.64$ & $43.1 \pm 12.4$  & $<0.01$ & $<0.01$ \\ \hline
                100 & $\textbf{2.81} \pm \textbf{2.23}$ & $4.70 \pm 4.49$ & $35.4 \pm 10.5$  & $<0.01$ & $<0.01$ \\ \hline
            \end{tabular}
            \caption{Average and standard deviation of disruptions across scheduling methods when flights are delayed (lower values are better). Scheduling density of 2. Buffer IP did not build a single schedule for 90 minutes and timed out, so we do not include it.}
            \label{dense_results}
        \end{table*}

    \subsection{Baseline Scheduling Performance}

        To show the efficacy of \name scheduling, we ran our best scheduling combination on the same disruption scenario, this time using different percentages $f$ of flights delayed. We selected $f$ values of 25\%, 50\%, 75\%, and 100\% and ran each scenario 100 times. We compared \name to directly using our RL agent or using the buffer integer program (with $T_{\text{buffer}} = 4 \text{ days}$). We show the results of our buffer-rewarded \name scheduler in Table \ref{disrupted_results_pm}. During these runs, the buffer IP and \name scheduler had similar run times, both averaging less than 0.85 seconds to create each schedule.

        We note that, because the constraints are exactly the same between the IP and \name scheduling approaches, we skipped recording the disruption value for the IP scheduler if and only if we also skipped the disruption value for the \name scheduler. Because of this alignment, the $p$-values comparing the two methods were obtained using a 2-tailed dependent t-test for paired samples between the \name and IP schedulers. This was not the case for the RL approach, which can generate a partial schedule, stopping when it is not able to assign a pilot to the next slot due to its previous decisions. When the RL approach created a partial schedule, we did not record its performance on that schedule to include in the average because it was unable to produce a full schedule like the baseline IP and \name schedulers. For this reason, in the \name vs. RL comparison, we use Welch's t-test for independent samples.

    \subsection{Highly-Constrained Scheduling Scenarios}

        In the schedule disruption scenario that we considered, the buffer IP formulation had little trouble building a robust schedule in a reasonable amount of time. However, the time advantages of \name become apparent in a more constrained scheduling setting. To demonstrate this efficiency, we performed the same experiment on \name, averaging over 100 trials. This time, though, we used a scheduling density of 2, creating twice the number of flights on average in each round of scheduling. This scenario is realistic in settings where, due to outside factors, many flights must be filled. For timing reasons, we only compared \name with the baseline IP and the pure RL scheduler. We show the results for this experiment in Table \ref{dense_results}. Importantly, across all flight delay percentages, the \name scheduler took an average of between 1.85 and 1.90 seconds to build a schedule, with a standard deviation between 0.55 and 0.60 seconds. We ran the exact same experiment for each disruption percentage for 10 iterations with the buffer-optimizing IP. However, we ended each experiment after 90 minutes, at which point the buffer IP had not finished building a single schedule. 

\section{Discussion}

    Our results clearly show the advantages of the \name approach. In the baseline scheduling scenario, \name produced schedules that provided 40\% to 45\% of the disruption reduction of the buffer IP compared to the baseline IP. In a powerful display of its usefulness, in a dense scheduling environment, \name performed 33\% to 48\% better than the baseline IP, producing 100 schedules with an average time of less than 2 seconds while the buffer IP failed to produce a single schedule in 90 minutes. In all of these experiments, the \name scheduler overwhelmingly outperformed the RL scheduler from which it was derived. These outcomes indicate that \name can harness various advantages of IP and RL scheduling to build a hybrid approach that improves on both methods used independently. 

    In the less-constrained baseline scheduling environment, \name performed worse than the buffer IP, though it still did better than the baseline. In this scenario, in a similar amount of time as \name, the buffer IP produced perfect schedules with no disruptions after flights were delayed. This result highlights the ideal use-case for \name: situations where approximate results are useful and the size of the integer program makes it infeasible to solve in a reasonable amount of time. The more constrained (density = 2) scenario fits this description well; the high number of flights in a shorter interval created more constraints and variables in the buffer formulation than our  IP solver could handle. By contrast, \name was able to produce a robust schedule in under 2 seconds, on average. 

    \subsection{Conclusions and Future Work}

    We introduced \name, a novel method for incorporating knowledge gleaned from reinforcement learning into an integer programming formulation. We applied this technique to a robust crew scheduling problem, looking at the assignment of pilots to flights so as to minimize schedule disruptions due to flight delays. We used \name to build a scheduler for this problem where the RL agent proposes weights for the selection of crew members, and the IP assigns the crew members using those weights. 
    In our experiments, \name outperformed both the baseline IP and the RL scheduler in creating schedules that are resistant to disruptions. Furthermore, in certain practical environments that caused the robust scheduling (buffer IP) formulation to be prohibitively slow, \name was able to create a robust schedule in a matter of seconds. 

    The introduction of nonlinear objectives or constraints can deteriorate the computational performance of MIP and IP solvers\footnote{The underlying formulations are no longer integer linear programs or mixed-integer linear programs}. However, the reward structure used to train an RL agent is not bound by such restrictions. While this paper used \name to approximate linear constraints and additional variables in an IP, it would be interesting to see how the approach performs when faced with nonlinear constraints.
     
     Finally, we have shown the efficacy of \name in robust crew scheduling. Given that IPs have long been a mainstay of discrete optimization, we believe that the approach could be useful in addressing other scheduling problems. Understanding the types of optimization problems for which \name is most effective is an interesting topic for further research.

\section*{Acknowledgements}
    The authors would like to thank the MIT SuperCloud \cite{supercloud} and the Lincoln Laboratory Supercomputing Center for providing high performance computing resources that have contributed to the research results reported within this paper. Research was sponsored by the United States Air Force Research Laboratory and the United States Air Force Artificial Intelligence Accelerator and was accomplished under Cooperative Agreement Number FA8750-19-2-1000. The views and conclusions contained in this document are those of the authors and should not be interpreted as representing the official policies, either expressed or implied, of the United States Air Force or the U.S. Government. The U.S. Government is authorized to reproduce and distribute reprints for Government purposes notwithstanding any copyright notion herein.

\bibliography{aaai22}

\clearpage
\appendix
\section{Appendix}
\subsection{Move-up Crews}

During our experimentation, we considered another factor that can lead to more robust schedules, move-up crews \cite{moveup}. We found that move-up crews, even when optimally scheduled, did not create particularly robust schedules, but we include our experimental results here for reference. For the sake of clarity, because we dealt with individual pilots rather than crews, we will depart from the literature and use the term move-up \textit{pilot} rather than move-up crew. A move-up pilot is someone who, by nature of their qualification and one of their scheduled flights, is readily available to move up to another flight should someone on that flight become unavailable, perhaps due to a delayed flight.

\subsubsection{IP Formulation}

We created an IP to increase move-up pilots in our schedules. We first define a threshold, $T_{\text{move}}$, for how far out we should look for move-up pilots. Now, we give a more formal definition of a move-up pilot: pilot $j \in I$, assigned to flight $g \in F$, is a move-up pilot for slot $s \in S_f$ on flight $f \in F$ if and only if all of the following conditions hold:

\begin{enumerate}
    \item $f \neq g$
    \item Flight $g$ starts at the same time as or later than flight $f$, and no later than $T_{\text{move}}$ days after $f$ starts.
    \item Flight $g$ ends at the same time as or later than flight $f$.
    \item Pilot $j$ is not on leave that overlaps with f.
    \item Pilot $j$ is not scheduled to any flights that start before $f$ and overlap with $f$.
    \item Pilot $j$ is qualified for slot $s$.
\end{enumerate}

Using constraints, we define the binary decision variable $M_{jg,fs}$ to be 1 if pilot $j$ assigned to flight $g$ is a move-up pilot for slot $s$ on flight $f$. We then use additional variables to build an objective function to maximize the number of move-up pilots in our final schedule. We can achieve this with the following objective function:

\begin{equation}
    \max \sum_{i \in I} \sum_{f, g \in F \times F} \sum_{s \in S_f} M_{jg,fs}
\end{equation}

\subsubsection{RL Training} 
    To train an RL scheduler to optimize for move-up pilots, we followed the same experimental procedure for the buffer-optimized RL scheduler, but we changed the reward function. For move-up pilots, we give a reward of $m + 1$ whenever our agent assigns pilot $j$ to a slot on flight $g$, where $m$ is the number slots on other flights that pilot $j$ can serve as a move-up pilot for. We use a maximum move-up time (like $T_{\text{move}}$) of 2 days. We did not do so out of concern for program run time; on a practical level, moving a pilot to a flight any more than 2 days earlier would cause a significant burden on the pilot rather than supply the convenient schedule alleviation that move-up pilots are supposed to provide. Just like the reward function for buffers, we include a -10 penalty for incomplete schedules and a +25 reward for complete schedules. We trained 15 total neural networks with the move-up pilot reward structure, using 5 different random seeds and schedule densities of 1, 2, and 3.

\subsubsection{Model Selection}

    We followed the same procedure as the buffer-rewarded networks to select the best network and $n$ value to use for our weight extraction method. We used $n$ values of 0, 2, 4, and 8. We used this procedure to obtain $r$, the ratio of average disruptions in the baseline IP schedule to average disruptions in the \name schedule with the chosen parameters. $r < 1$ indicates that the \name schedule had fewer disruptions on average. The best median $r$ value was 0.46, with a range of 0.38 (0.25 to 0.63), produced with a scheduling density of 2 and an $n$ value of 2. Two networks with these parameters and different seed values produced the median value, so we picked one arbitrarily. We used this model in our subsequent experiments. 

    The range across seeds for our move-up-rewarded \name scheduler (0.38) was notably higher than the range for our best buffer-rewarded scheduling method, which was 0.08. Like the buffer-rewarded schedulers, the $n$/density combinations also had fairly stable performance across random seeds. The 3 highest ranges across random seeds were 0.88, 0.69, and 0.68. Similar to the buffer-rewarded schedulers, the 2 highest ranges were produced by the ``blank slate'' weight extraction method ($n = 0$).

\begin{table*}[ht]
\centering
\begin{tabular}{|l|l|l|l|l|l|l|l|}
\hline
{\% Flights} & \multicolumn{4}{c|}{Number of Disruptions} & \multicolumn{3}{c|}{Significance of Difference} \\ 
 & \multicolumn{4}{l|}{} & \multicolumn{3}{c|}{($p$-value)} \\ \cline{2-8} 
Delayed   & \multicolumn{1}{c|}{\name}           & \multicolumn{1}{c|}{Baseline IP}     & \multicolumn{1}{c|}{RL}              & \multicolumn{1}{c|}{Buffer IP}       & \name-& \name-  & \name-        \\ 
    & & & & & Baseline & RL  & Buffer        \\  \hline
25                                  & $\textbf{0.63}\pm \textbf{0.93}$  & $0.61 \pm 1.07$ & $32.6 \pm 7.42$ & $0.33 \pm 0.63$ & .88       & $<.01$   & $<.01$             \\ \hline
50                                  & $\textbf{1.05} \pm \textbf{1.13}$ & $1.16 \pm 1.55$ & $27.6 \pm 7.24$ & $0.68 \pm 0.90$ &  .51   & $<.01$  & $<.01$        \\ \hline
75                                  & $\textbf{1.19} \pm \textbf{1.25}$ & $1.13 \pm 1.73$ & $23.9 \pm 6.44$ & $0.74 \pm 1.04$ & .73        & $<.01$  &  $<.01$        \\ \hline
100                                 & $\textbf{1.11} \pm \textbf{1.14}$ & $1.06 \pm 1.50$ & $19.1 \pm 6.47$ & $0.65 \pm 0.88$   & .76    & $<.01$  & $<.01$            \\ \hline
\end{tabular}
\caption{Average and standard deviation of disruptions across scheduling methods when flights are delayed (lower the better). Scheduling density of 1. The \name and RL schedulers used the move-up reward function in their underlying neural network.}
\label{disrupted_results_moveup}
\end{table*}

\subsubsection{Baseline Scheduling Performance}

    Using the selected model and $n$ value, we ran the same disruption scenario as the buffer-rewarded \name scheduler over 100 iterations. We compared the \name scheduler against the baseline IP scheduler, the RL scheduler with the same underlying neural network, and the move-up IP scheduler with a $T_{\text{move}}$ value of 2. The results are shown in Table \ref{disrupted_results_moveup}.

\subsubsection{Discussion}

    Based on our results, the move-up IP scheduler produced more robust schedules than any of the other methods, but it did not perform as strongly as the buffer IP scheduler, which entirely eliminated schedule disruptions across all percentages of flights delayed in our previous experiment. This weakness likely explains the performance of the \name scheduler based on the move-up reward function, which showed no significant difference in schedule disruptions compared to the baseline IP scheduler. Because of the relative inefficacy of increasing the number of move-up pilots in producing robust schedules, we decided to focus our efforts on using buffers to reduce disruptions in our main work.

\end{document}